# Do we need Asimov's Laws?

Ulrike Barthelmess, Koblenz
Ulrich Furbach, University Koblenz

**Abstract:**

In this essay the stance on robots is discussed. The attitude against robots in history, starting in Ancient Greek culture until the industrial revolution is described. The uncanny valley and some possible explanations are given. Some differences in Western and Asian understanding of robots are listed and finally we answer the question raised with the title.

Robots and autonomous machines are very often central characters in science fiction literature and movies. A very common plot in this regard are machines that are going crazy or even taking over the rule about mankind. This essay is dealing with the question, why robots are characterized like this so frequently and we will discuss that this was not always the case in history.

The notion 'robot' goes back to a play written by Karel Čapek in 1924 ([ČN04]). Robota is the Czech word for forced labour and it became introduced into English and many other languages by this play. Already here, so to speak, in the cradle of robotic culture, the robots took over the world and nearly destroyed mankind. One of the best known science fiction authors, Peter Asimov, did not like the usual 'Frankenstein' pattern: a robot was created and at the end destroyed by his creator[1] – he explained this in the introduction of [Asi64] in 1964. This is why he introduced his famous three laws of robotics in [Asi42]:

1. A robot may not injure a human being or, through inaction, allow a human being to come to harm.

2. A robot must obey the orders given to it by human beings, except where such orders would conflict with the First Law.

3. A robot must protect its own existence as long as such protection does not conflict with the First or Second Laws.

---

[1] We will see later on that we should better call this the 'Golem' pattern

Asimov wanted to overcome the widespread robot Frankenstein-like plots in science fiction literature by making it unnecessary to destroy the robot in order to prevent him from harming man or even mankind. Although this motivation is sound, it implies that robots are dangerous in general – the creator of the robot have to implement these laws in order to protect himself. What is a reason for this fear of robots going crazy?

In the following Section 1 we will discuss some developments of artificial beings during various epochs of history. In Section 2 we will focus on aspects of our cultural background and in Section 3 we will introduce the uncanny valley and discuss its possible reasons and finally we investigate cultural differences.

1. **Robots in History**

In this section we discuss some robotic development from far before Asimov and we will see they were all introduced without the need of protecting man by specific laws. Many more examples from culture and art can be found in [BF11].

When asked about early historical robots most people mention immediately some ma- chines designed by Leonardo da Vinci; and, indeed, he specified and designed a whole variety of mechanical devices, which deserve to be called 'robots'[2]. He designed a programmable autonomous vehicle and besides many military vehicles and machines, there are also plans for an android robot and a moving and roaring lion designed by da Vinci ([Ros06]).

However, there are examples about artificial live much earlier in history of culture. The first ones we want to mention were very useful robots in action; in fact it is a very early mention of an assistant system, just the way they are common nowadays in cars or air- planes. It was Homer who described in the 18th book of the Iliad, how the mother of Achilles, Thetis, visits the divine blacksmith Hephaestus in order to get a new strong suit of armour for her son. Homer describes the workshop and the moving around of the lame Hephaestus:

> There were golden handmaids also who worked for him, and
> were like real young women, with sense and reason, voice also

---

[2] In the following we use the word robot, even for machines which were constructed or envisioned long before the term robot was introduced

and strength, and all the learning of the immortals; these busied themselves as the king bade them, ···

Another robot story is that of Galatea and Pygmalion. According to Greek mythology Pygmalion was an artist from Cyprius who carved a woman out of ivory and fell in love with her. After prayers to god Aphrodite, the statue, Galatea, became alive; she lived together with Pygmalion, they had a son and according to some sources also a daughter. The topic of Galatea and Pygmalion inspired many artists: In George Bernhard Shaw's 'Pygmalion' the flower girl Eliza Doolittle is in some sense created by Mr. Higgins. In E. T. A. Hoffman's 'Sandmann' a young man falls in love with Olimpia, an automata; this is very nicely transferred in Jacque Offenbach's opera 'The Tales of Hoffmann', in which during 'The doll song' the automata Olimpia has to be wound up several times. But Hoffmann does not notice that he is blinded by love with this automata.

There is a parallel aspect of this blindness when we have a look at the Mechanical Turk, constructed by Manfred von Kempelen in 1770. This was a fake chess playing machine which impressed for many decades the old and the new world. Maria Theresia of Austria, Napoleon Bonaparte and Benjamin Franklin, among many others, admired this 'machine'. There have been many books published about this Turk and it was touring through Europe and North America until it was destroyed by a fire in 1854. In the 1820s it was exposed as a fake machine – notice that it took more than 50 years until it was recognized as a hoax. Our conjecture is – to come back to the blindness of Olimpias lover – that people of that time wanted to believe. This appears not too astonishing, if we take into account that during the age of enlightening the focus in philosophy, science and even medicine was on rationality and the use of a scientific method. In Germany it was Immanuel Kant who used the word 'Aufklärung' in order to name this development. In a seminal essay he cited sapere aude – dare to know – which became a kind of leitmotif for the age of enlightenment. Gottfried Wilhelm Leibniz, who was a mathematician and philosopher, developed the infinitesimal calculus. He constructed mechanical computers and he was dreaming about mechanical reasoning. He was convinced that human reasoning can be reduced to symbolic calculation and instead debating about different opinions in the future one would sit down and say calculemus!

In this climate the development of constructing automata and robots, which already started during Renaissance and Baroque, was flourishing. There were all kinds of automata and even android robots that could be programmed in order to write letters, produce drawings or play music. And, of course, the already discussed mechanical Turk could celebrate its success during this period.

Altogether, the examples we gave from various episodes of history of culture do not give any hints that the inventors or constructors had any fear of danger which might come from these automata robots.

## 2 The Dark Side of Robots

In this section we will describe two aspects which might explain, why robot and automata stories very often come with a 'Frankenstein-plot'. But first of all a note on Frankenstein: Mary Shelley published her novella about Dr. Frankenstein who created a man that became a monster in 1818. In contrast to various famous film adaptions the creature in the original book became a monster not because of some errors during the creation process – it became evil, murdered and turned out to be disaster for his creator, because it was not accepted as a human being. The monster wanted Frankenstein to build a partner for him who did not meet this desire, and this was the reason for all that harm. It was not the case that the monster was evil and violent from the very beginning, it became like this by not being accepted as a social and emotional being.

A very characteristic saga for the more negative consequences of creating life is the Golem narrative. It occurs in various epochs and in different versions – the most common one might be the story of Rabbi Löw from Prague. This was in the late 16th century and the Jews in Prague were living in a ghetto, where they suffered from antisemitic attacks. The Rabbi, following a dream he had, constructed the Golem out of clay and after some special hocus-pocus the golem came to live. He did his job to protect the Jewish community perfectly well. He had to be deactivated on Sabbath; once the Rabbi forgot this deactivation and in most versions of the story the golem started to become a monster and had to be destroyed.

**Religion** There are versions of the Golem narrative which go back to the Tora and several myths. The very obvious explanation for the failure of the Golem project is the fact that in most Christian religions it is forbidden for man to be godlike. There are numerous examples for breaches of this regulation in the Bible; e.g. already in the book of Genesis 2:24 the serpent was talking to the woman 'For God knows that when you eat of it your eyes will be opened, and you will be like God, knowing good and evil.' And, of course, this attempt to be godlike was punished. Another example is the Tower of Babylon, where people wanted to construct a building which could reach the sky. And God came down and he wanted to prevent people from reaching their goal because if they would succeed as one people with one language, 'nothing they plan to do, will be impossible for them' (Genesis 11:1). This is why God confused their language, such they were not able to understand each other and thus the construction of the tower failed.

In both examples we see the pattern which also applies to the Golem narrative: if we try to be godlike by creating live, we get punished. This pattern is repeated by numerous science fiction stories and also in several versions of the Frankenstein story. Our hypothesis is that this pattern is based on our Christian religious cultural history. We do not elaborate this further, but it should be obvious that this is not a mere Christian feature, moreover it is similar in all Abrahamic religions. And moreover, a similar aspect can be found in many ancient Greek tragedies, namely the concept of hubris. Many figures in Greek antiquity, like Prometheus or Niobe, are punished in the end, because the act with arrogance towards the Gods.

**Industrial Revolution** The second aspect which might be responsible for the negative feelings which come up very often when robots enter our daily environment is the experience of the industrial revolution in Europe. Certainly, the age of enlightenment, which we discussed before as a promoter of the robotic developments, can be seen as the pathfinder of industrialization in the late 18th and the 19th century. But note that we are talking about the way industrialization spread over Europe and influenced every part of daily life. It was a revolution which turned the living conditions of large parts of the population. There was a significant rural exodus, cities grew larger and at the same time the living and working conditions of the working class changed dramatically to the worse. The technical development in engineering, mining, chemistry and transportation resulted in a dominance of machines, which often was felt as a threat. There was a kind of technophobia which even resulted in fights against machines. In England there was a movement which destroyed weaving machines which became so grave that the Parliament decided demolition of ma- chines to be a capital crime. One group, called the Luddites, even fought battles against the British army; in the beginning of the 19th century there were a number of show trials against these machine destroyers, which even resulted in a number of executions. At that time machines began to dominate large groups of workers. A very imposing treatment of this development is the awesome movie 'Modern Times' by Charlie Chaplin.

We will come back to the arguments we gave in this section, when we will discuss the differences with Asian countries, like e.g. Japan.

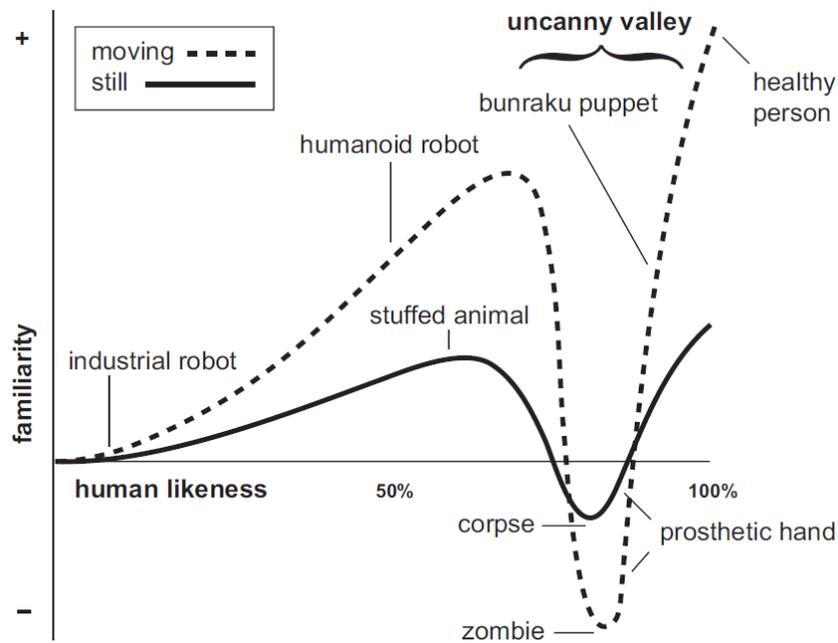

Figure 1: The Uncanny Valley ([Mac05a])

## 3 Uncanny Valley

The phenomenon of the uncanny valley was introduced by Masahiro Mori, a Japanese robot designer, in 1970 (cited from [Mac05a]) as a description of how people feel when confronted with robots. The graphic in Figure 1 depicts the familiarity we feel when confronted with a robot, depending on its human likeness. At the beginning the graph increases, the more human-like the robot is - but in a stage of increasing human likeness, familiarity is decreasing suddenly. Only after the robot is getting much more humanlike we feel positive again. This effect is even more significant if the robot is able to move. Nowadays the uncanny valley is also of importance for the movie industry. There is a number of films, in which reviewers reported a kind of revulsion when looking to computer animated figures, which looked too realistic. In the meantime there is even a design methodology on how to avoid the uncanny valley in animated movies.

A possible explanation for this behavior is discussed in [Mac05b], namely the terror management theory. This is a theory from social psychology which basically forms the hypothesis that humans are trying to give their lives a certain meaning, in particular because they are aware of death. However, death creates an anxiety that they are trying to avoid. In [GPS$^+$90] the authors give a series of experiments which show clearly that reminding people of their mortality increases attraction to those who consensually validate their belief (because they origin from the same culture, religion etc.); it decreases attraction to those who threaten their believes. Robots which are on the rim of being fully human-like may remind people of their mortality and they clearly are not from the same social group; this is why in the uncanny valley familiarity goes down when robots remind us on our mortality. The same mechanism might also explain why we usually regard zombies as creepy.

Besides the terror management theory explanation there is another theory which might help explaining the uncanny valley, namely the category boundary problem. According to [MVH09][3] people feel disturbed about things that cross category boundaries. In many cultures ambisexual people are forced to decide themselves for one category, male or female, the border is not accepted. Another example is one of the dietary laws from the bible (Book of Leviticus 11), where food which does not fit into categories is restricted. Pork is forbidden because pigs are cloven-hoofed, but unlike other cloven-hoofed animals they do not chew the cud. Last not least we should mention zombies as an example. Zombies are very much like humans, but they are not alive; they cross a boundary. With robots the situation is even more drastic: they can be android, having a lot of similarities with humans, they are constructed by humans, and they are not alive. They exist on the boundary between two classes, and even worse, we are belonging to one of it.

## 4 Cultural Differences

We started this essay with the argument that it is a very common plot of science fiction stories that robots get crazy and threaten humans or even mankind. In the previous sections we gave some possible explanations of this aloofness against robots; some of the arguments have been based on religious aspects. Although we have been using arguments from the bible, it is obvious that similar reasoning holds for all Abrahamic religions.

---

[3] The argumentation of MacDorman is based on various other publications about android science

Looking to Asia, in particular Japan, we find a very different stance on robots. For example one of the most famous heroes from mangas is Astro Boy, a robot boy which definitely is a good robot. In 2007 Astro Boy was named by the Ministry of Foreign Affairs as the Japanese envoy for safe overseas travel. Furthermore, there is even a Japanese city which lists Astro Boy as a registered citizen. Robots are very present in Japanese culture and research. In research the most noticeable aspect is that there is a lot of work towards robots as a means for communication. Humanoid robots are used for experiments on social interaction with kids in elementary schools and the probably best known android scientist, Hiroshi Ishiguro, is constructing humanoids, among them a copy of himself, which can replace the professor in meetings. All that would be regarded a bit unusual in Western universities.

According to the MIT historian Bruce Mazlish one reason for this difference are certain discontinuities of western culture ([Maz95]). There are four of these discontinuities: the earth as the center of our universe, the creation myth of Genesis, the rationality assumption about mind from the age of enlightening and finally the separation of body and mind. During western history of culture the first three of these discontinuities have been demolished: the Copernican Revolution, Darwin's theory of selection and Freud's work on unconsciousness. The destruction of these mind-settings influenced our self-confidence significantly; it undermined our personal and human identity.

According to Mazlish, the forth discontinuity is on the way to be destroyed by the spread of robots - and this is threatening! We are tending towards renegading this development. The differences concerning the stance on robots between Western and Asian culture might stem from the fact that none of the above depicted demolitions of discontinuities occurred in Eastern culture. They are missing the negative experience of these dramatical turns in the various fields of culture and science. And hence there is no fear of another destruction of central believes. A very similar argumentation results from looking at the industrialization argument we gave before. Although there was an industrialization in Japan, in particular after World War II, it was not a revolution as it happened in Europe. Until the 1950th Japan had a rather low level of development and this is why industrialization caused a spectacular growth of economics and the same time an improving of living conditions. There was no need to fight against machines as the Luddists in England did a century before.

Another explanation for the difference in Eastern culture is the religion argument. Before, we showed that in Christian and in general in Abrahamistic religions it is forbidden to be godlike, every attempt to act like God, e.g. by creating life (e.g. Golem) or reaching heaven (Babylon) was punished. There is no such rule in Eastern religion, in particular in traditional Shintoism, which is an animistic religion, claiming that everything has a spiritual essence. This even holds for inanimate things, and hence for robots.

Makoto Nishimura, a Japanese pioneer in robotic, put it already in 1928 the following way: 'If humans are the children of nature, than artificial humans are the grandchildren of nature'.

There are empirical investigations in android science concerning the differences between Western and Eastern culture. E.g. MacDorman did a series of experiments with Japanese and American university faculties ([MVH09]). Unfortunately it was not possible to prove any differences - but the reason could be in the selection of highly educated faculties as test persons.

## 5  Answering the Title Question

In the previous sections we gave a series of explanations for the stance on robots, which can be observed in Western culture. Asimov's laws are a kind of attempt to protect humans against robots. We want to argue that there is no real menace by robots, the fear stems from the history of culture and from religious traditions. To answer the initial question of this paper:

> There is no need of Asimov's laws!

At least there is no need to teach these rules to our robots. Sometimes they are designed in a way, that Asimov's laws are implicit consequences of their functionality. As an example you may take the development of modern cars. They are turning more and more into a kind of autonomous vehicles: They have track assistants, they automatically detect obstacles and start braking and they even navigate into parking slots. All that is obviously designed aiming at road accident prevention and as such it is a means to protect humans. No one ever claimed that Asimov's laws have to be implemented in those systems.[4]  The same holds for airplanes, which nowadays fly by wire which actually means, that a lot of flying activities are not done by pilots but by computers.

---

[4] In Nevada self-driving cars even get licensed and hit the streets

But we also should mention that there do exist autonomous vehicles and robots designed per se to harm humans. Military robots or autonomous drones are aiming explicitly at violating Asimov's laws. What we desperately need are legal and ethical rules for the commitment of robots. We can see this from the debate around drone strikes in Pakistan, Yemen and Somali. According to the Bureau of Investigative Journalism[5] there is a kind of covert drone war in those countries. Drones are used to strike against targets in countries, without being officially in war according to the international law of armed conflict. More or less autonomously operating drones are destroying targets, i.e. humans, which are associated with terrorism. And as can easily be imagined there is a significant number of civilians killed or injured as 'collateral damage'. We want to argue that a similar procedure would not so readily be accepted by the world public, if instead of drones manned aircrafts would be used. It seems as if there is much lower acceptance threshold to use robots instead of regular military forces for illegal or covert ware fare. Besides of moral and ethical considerations, there also are a lot of legal questions. Is it legal to strike targets with unmanned drones in a country which is not in a formal state of war with the owner of drones? Is it legal for a third party country to support such an action, e.g. by delivering data for military reconnaissance or by hosting the 'pilots' of the drones?

In the context of this discussion it would be more likely to answer the question from the title as follows:

> It is not allowed to build and to use robots which violate Asimov's first law!

As an example of such a law we could use the Chemical Weapons Convention (CWC), a treaty which prohibits the use and production of chemical weapons, as well as the destruction of all chemical weapons. We are aware that it would be very difficult to keep such a ban under surveillance. It is easy to arm a previously unoffending drone and turn it into a weapon which would be covered by the ban.

In this paper we discussed problems concerning the harmonization between humans and robots. But maybe all that will not an issue anymore, if we believe Ray Kurzweil. He is forecasting that humans and robots will reach a point, the so called singularity, where we humans fuse with robots. He claims that by the year 2045 technology will reach a point, where humans are not able to comprehend it. Moreover, they will augment their minds and their bodies by artificial means and parts. What will be the difference of humans augmented by technology and artificial intelligence and robots?

---

[5] http://www.thebureauinvestigates.com